\documentclass[10pt,final,twocolumn]{IEEEtran}
\usepackage{amsmath}
\usepackage{latexsym}
\usepackage{amssymb}
\usepackage{graphics}
\usepackage{graphicx}
\usepackage{epsfig}
\usepackage{booktabs}
\usepackage{dsfont}
\usepackage{subfigure}

\usepackage[ruled,vlined]{algorithm2e}
\usepackage{algorithmic}
\usepackage{amsfonts}
\usepackage{multirow}
\usepackage{multicol}
\usepackage{xcolor,colortbl}
\usepackage{amsthm}
\usepackage{amssymb}

\newcolumntype{a}{>{\columncolor{Gray}}c}
\makeatletter
\def\maximize{\mathop{\operator@font maximize}\limits}
\def\minimize{\mathop{\operator@font minimize}\limits}
\makeatother

\definecolor{Gray}{gray}{0.9}

\usepackage{mathtools}

\newcommand{\be}{\begin{enumerate}}
\newcommand{\ee}{\end{enumerate}}

\def\maximize{\mathop{\operator@font maximize}\limits}
\def\minimize{\mathop{\operator@font minimize}\limits}
\makeatother

\begin{document}

\title{An Interpretable Compression and Classification System: Theory and Applications}
\author{Tzu-Wei Tseng, Kai-Jiun Yang, C.-C. Jay Kuo and Shang-Ho (Lawrence) Tsai$^{\dagger}$}

\maketitle

\begin{abstract}
This study proposes a low-complexity interpretable classification system.
The proposed system contains  main modules including feature extraction, feature reduction, and classification.
All of them are linear.
Thanks to the linear property, the extracted and reduced features can be inversed to original data, like a linear transform such as Fourier transform, so that one can quantify and visualize the contribution of individual features towards the original data.
Also, the reduced features and reversibility naturally endure the proposed system ability of data compression.
This system can significantly compress data with a small percent deviation between the compressed and the original data.
At the same time, when the compressed data is used for classification, it still achieves high testing accuracy.
Furthermore, we observe that the extracted features of the proposed system can be approximated to uncorrelated Gaussian random variables.
Hence, classical theory in estimation and detection can be applied for classification.
This motivates us to propose using a MAP (maximum a posteriori) based classification method.
As a result, the extracted features and the corresponding performance have statistical meaning and mathematically interpretable.
Simulation results show that the proposed classification system not only enjoys significant reduced training and testing time but also high testing accuracy compared to the conventional schemes.
\end{abstract}

\footnotetext[1]{S.-H. Tsai and T.-W. Tseng are with the department of Electrical Engineering, National Chiao Tung University, Hsinchu, Taiwan.  E-mails: shanghot@alumni.usc.edu and  ryan8314brad@gmail.com. This research was supported by the Ministry of Science and Technology (MOST), Taiwan under Grant MOST 107-2221-E-009-065. }
\footnotetext[2]{K.-J. Yang is with the Industrial Technology Research Institute, Zhudong 31057, Taiwan.  E-mail: kaijiuny.ece98g@g2.nctu.edu.tw.}
\footnotetext[3]{C.-C. J. Kuo is with the department of Electrical Engineering, Signal and Image Processing Institute, University of Southern California, Los Angeles, USA.  E-mail:  cckuo@sipi.usc.edu.}


\section{Introduction}

Classification for multimedia has been studied and applied to a variety of applications such as security, entertainment, and forensics for years.
The development of intelligent algorithms for classification results in efficiency improvements, innovations, and cost savings in several areas.
However, classification based on visual content is a challenging task because there is usually a large amount of intra-class variability, caused by different lighting conditions, misalignment, blur, and occlusion.
To overcome the difficulty, numerous feature extraction modules have been studied to extract the most significant features and the developed models can achieve higher accuracy.
Selecting a suitable classification module to handle the extracted features according to their properties is also important.
A good combination of feature extraction and classification modules is the key to attain high accuracy in image classification problems.

In general, classification models can be divided into two categories: 1) convolutional neural networks (CNN) structure and 2) non-CNN based structure.
The CNN structures are in the mainstream \cite{facenet}-\cite{resnet}, and they are usually stacked up with several convolution layers, max-pooling layers, and ReLU layers.
Hence the depth of the structure can be deep such as ResNet \cite{resnet}.
Due to the deep structure, CNN models can extract the features well and achieve high accuracies.
However, CNN models usually share several disadvantages, including mathematically intractable, irreversible, and time-consuming.
Most of the CNN models use handcrafted-based feature extraction or backpropagation to train the convolution layers which makes the models mathematically inexplicable and time-consuming. In addition, the max pooling and ReLU layers are nonlinear, hence the structures become irreversible.
The above disadvantages increase the difficulty in designing and adjusting suitable structures for various types of data sources.
Therefore, research has been conducted to design a model which can be easily and fully interpretable mathematically like the second category introduced below.

Classification models using non-CNN based structures are mainly built with two parts, feature extraction modules, and classification modules.
To design a model which is mathematically interpretable, several machine learning techniques for feature extraction or classification have been developed including various versions of principal component analysis (PCA) \cite{saak}-\cite{incremental}, linear discriminant analysis (LDA) \cite{linear}-\cite{human}, support vector machine (SVM), nearest neighbor (NN), etc.
For the feature extraction modules, in \cite{saak}-\cite{incremental} the authors used PCA; while in \cite{linear}-\cite{human}, the authors used LDA to extract features from images.
For the classification modules, \cite{pcanet}, \cite{face}, \cite{feature}, \cite{incremental} and \cite{palmprint} applied the nearest neighbor classifier.
On the other hand, \cite{saak} and \cite{comparative} utilized the SVM classifier.
The combinations of feature extraction and classification modules are generally nonlinear.
Thus to inverse or recover the extracted features to original data is difficult.

Inversing the features to original data reveals important information for classification.
For instance, if certain significant features are extracted and they result in high classification accuracy,
one would be interested in visualizing or quantifying these features in the original multimedia if this is feasible.
The reversibility is like Fourier series.
One can inverse individual Fourier series, quantify how they contribute to the original signals and understand the importance of individual series.
Due to the difficulty of data reversibility in most existing solutions, methods to compress data efficiently dedicated for classification purposes have not been well addressed yet.

Data compression for classification purposes leads to not only less storage but also lower computational complexity.
For instance, in a classification problem, if the extracted features can be reduced while the reduced features can achieve satisfactory testing accuracy and the inversed images from the reduced features still keep important characteristics of the original images, one may store the reduced features instead of all features.
Also, the reduced features can be used for classification to decrease computational complexity.
Furthermore, such data compression is interpretable because it keeps the most significant features for classifications.
It is worth mentioning that the concept ``compression'' here is different from that in the conventional compression techniques such as JPEG 2000 and HEVC intra coding.
Compression for classification aims to compress data as well as achieve high classification accuracy in verification stage.
On the other hand, the conventional compression methods attempt to maintain visual recognition when compressing data.

In this study, we propose an interpretable classification system that can achieve high testing accuracy with low complexity.
The proposed system is linear and invertible.
Hence data compression via the system is possible.
The proposed system can be divided into several parts including feature extraction, feature reduction and the classification for the reduced features.
The feature extraction and reduction modules consist of consecutive PCAs with truncated features, and the classification module detects the class of input images using the maximum a posteriori (MAP) detector.
Thanks to the linear property of the proposed system, the extracted features can be inversed to original data to see insight into individual features.
Thus, it is like a linear transform, such as Fourier transform, where both forward and backward directions are feasible.
The proposed system can also significantly reduce the extracted features while it still maintains high classification accuracy.
As a result, data compression for classification purposes is realizable via the proposed system.
Moreover, the computational complexity is significantly reduced due to the use of cascade PCAs with dimension reduction and MAP detector.
Thus both the training and verification time of the proposed system are only $1/100-1/1000$ of those using AlexNet and Saak transform.
Furthermore, we find that the extracted features in the proposed system can be approximated to uncorrelated Gaussian random variables.
It is worth pointing out that the Gaussian approximation result was also observed in \cite{saak}.
The Gaussian approximation result endues the proposed system and the extracted features statistical meaning.
Hence classical estimation and detection theory can be applied to the proposed system for classifying the data \cite{kay}.
This motivates us to use the concept of maximum a posteriori (MAP) to detect the class of input images.
Therefore for every input image, there is a probability that this image belongs to each class.
The detected class is the one that has the maximum probability.
Since every class has a probability for the input data, one can also determine top candidate classes for the data and develop more sophisticated algorithms to refine the classification results.
Consequently the proposal is an interpretable compression and classification system.

To verify the classification ability of the proposed system, we conduct experiments in  face recognition “who is this person?”.
We use the dataset in \cite{LFW}, Labeled Faces in the Wild (LFW). LFW is widely used by face recognition models like \cite{facenet}-\cite{robustfr} and \cite{pcanet}.
Experiment results show that the proposed scheme outperforms conventional systems in terms of both testing accuracy and computational complexity.
Moreover, the training and testing time of the proposed system is much faster than conventional schemes.
In a standard PC platform, two datasets with 2804 and 6592 images only take 0.7 and 1.7 seconds for training respectively.
Also, it takes less than 0.15 msec to recognize the class of one image.
The accuracy reaches 97.61\% for a 19-person dataset and 84.91\% for a 158-person dataset.
Furthermore, thanks to the linear property, the proposed system can inverse the reduced features to the original image and the compression ratio is significant.
In our experiments, the proposed system can reduce the number of features from 12288 to 270, a compression ratio up to 45.51:1; at the same time, the average deviation between the original and compressed images is only 9.11\%, and more importantly the
testing accuracy is 97.61\% for the 19-person dataset and 84.91\% for the 158-person dataset.

The outline of the remaining parts is organized as follow: In Section II, we present the linear recognition model and corresponding algorithms.
In Sections III-IV, individual modules of the proposed system are explained in details, where  Section~III explains the linear feature extraction module, and Section~IV introduces the linear classification module.
In Section V, we provide experimental results to show the advantages of the proposed system.
Conclusions and future works are given in Section VI.

\section{Proposed Recognition Model and Methods}\label{secprmm}
\begin{figure*}
    \centering
    \includegraphics[width=1\textwidth]{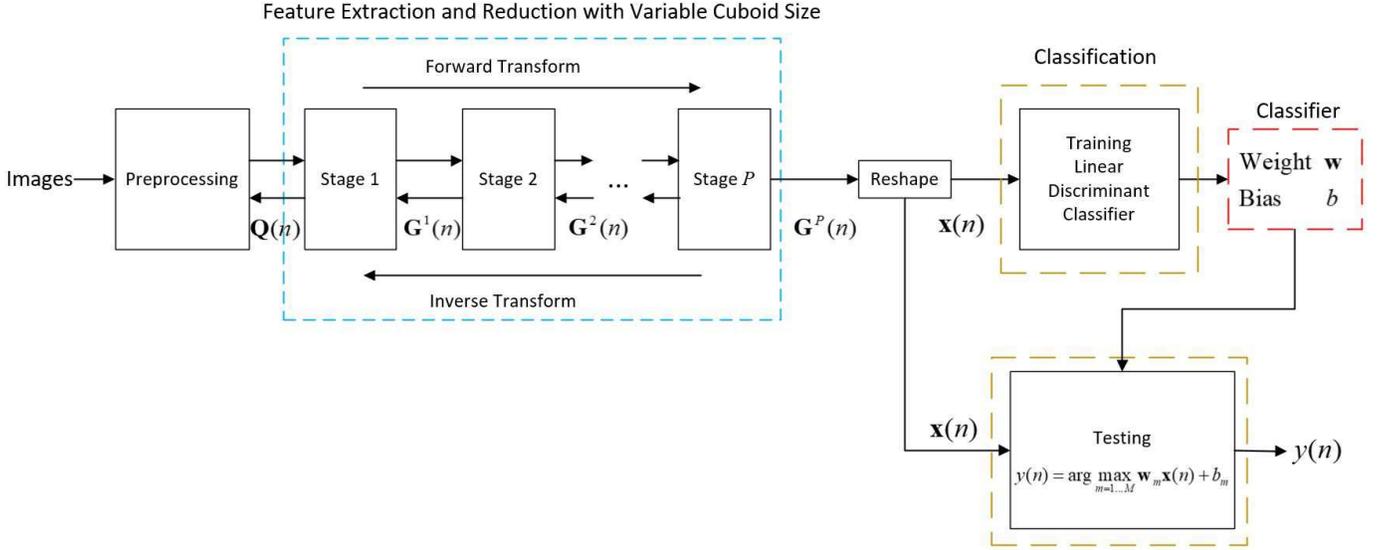}
    \caption{The  proposed linear classification system consisting of the preprocessing, feature extraction and reduction, and classification modules.}
    \label{flowchart}
\end{figure*}
A block diagram of the proposed recognition model is shown in Fig. \ref{flowchart}, which consists of preprocessing, feature extraction and reduction, and classification modules.
The preprocessing module handles the original defects of the dataset such as noise, contrast, and brightness of the images.
The preprocessing module contains operations including object detection, image processing, and data augmentation. It is an adjustable module designed according to the dataset.
The feature extraction module extracts the significant features out from the dataset and has multiple stages.
Each of the stages is a linear transformation.
Hence the transformation is invertible, and the images can be reconstructed from the extracted features.
This point will become more clear later.
After the feature extraction module, features are reduced for data compression purpose and avoiding overfitting.
Finally, the reduced features are passed through the classification module, where we propose to use linear discriminant classifier because we find that a Gaussian statistical model can be assumed after the proposed linear transformation.
Because the proposed system is linear and based on statistics, many results can be explained and improved mathematically.
Let us explain individual modules more detailed in the following sections.

\section{Feature extraction and reduction with variable cuboid size}\label{secfeature}

In this section, we introduce the feature extraction and reduction module, which is linear and hence both forward direction (data to extracted features) and inverse direction (features to original data) are feasible.
This module can be regarded as a consecutive PCA (Principal Component Analysis) with truncation in each stage.
The extracted features can be assumed to be uncorrelated Gaussian random variables.
Those points are explained in the following subsections.

\subsection{Forward Direction}

The proposed feature extraction module is a fast linear data-driven feedforward transformation with low space complexity.
Referring to the block diagram in
Fig. \ref{flowchart},
let $\mathbf{Q}(n)\in R^{I\times J}$ be the $n$th preprocessed image of a size of $I\times J$, where $n=1,2,\cdots,N$.
Hence $N$ is the total number of the input images.
In the testing phase, the value of $N$ can be one; while in the training phase, it has to be more than one to find the transform kernels.
Then, we feed all of the preprocessed images into the multi-stage proposed scheme and let the initial global cuboid $\mathbf{G}^0(n)\in R^{I^0\times J^0\times K^0}$ be $\mathbf{Q}(n)$, where the superscript is the stage index and
$I^0=I$,
$J^0=J$ and $K^0=1$. Here, $I$ and $J$ are the global cuboid sizes at the vertical and horizontal directions in the spatial plane respectively, and $K$ is the global cuboid size at the spectral direction.

\begin{figure}
    \centering
    \includegraphics[width=0.5\textwidth]{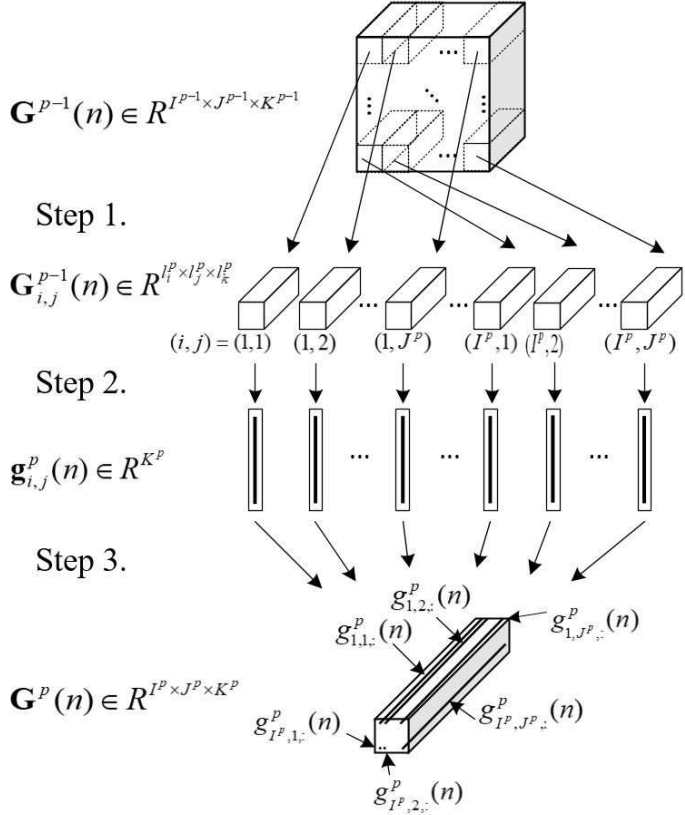}
    \caption{The dataflow, reshape operations and dimensions of cuboids  between Steps $p-1$ and $p$.}
    \label{stagep}
\end{figure}

Assume that there are $P$ stages.
At each of stage, we reshape the global cuboid into multiple non-overlapping local cuboids, perform principal component analysis on individual cuboids, collect the results, and reshape them into another global cuboid for the next stage.
According to the size of input images, one can adjust the side lengths of the local cuboids in the vertical and horizontal direction at each stage.
Let the side lengths of the local cuboids in stage $p$ be $l^p_i\times l^p_j\times l^p_k$, where $l^p_i$,  $l^p_j$ are adjustable. The values of $l^p_i$, $l^p_j$ and $l^p_k$ satisfy
\begin{align}
&\prod^P_{p=1}l^p_i=I,\;\;I^{p-1}/l^p_i\in N,\\
&\prod^P_{p=1}l^p_j=J,\;\;J^{p-1}/l^p_j\in N,\\
&l^p_k=K^{p-1}.
\end{align}
When the initial input and the side lengths $l^p_i$,  $l^p_j$ are given,
the input can be processed by the multi-stage scheme.
The dataflow and the dimension conversion at Stage $p$ are shown in Fig.~\ref{stagep} and also explained in Steps~1-3 below: The stage index $p$ is with increasing order, {\em i.e.,} $p=1, 2, \cdots P$.

\textbf{Step 1. Global cuboid to several local cuboids.}
Let the global cuboid of the $n$th image at Stage $p-1$ be $\mathbf{G}^{p-1}(n)$ with  size $I^{p-1}\times J^{p-1}\times K^{p-1}$.
The global cuboid is cut into several local cuboids $\mathbf{G}^{p-1}_{i,j}(n)$ with size $l^p_i\times l^p_j\times l^p_k$
at stage $p$.
With the constraints of the local cuboid side lengths, one can perfectly cut the global cuboid into $I^p\times J^p$ non-overlapping local cuboids below
\begin{align}\label{decomposition}
\mathbf{G}^{p-1}(n)=
\begin{bmatrix}
    \mathbf{G}^{p-1}_{1,1}(n) & \mathbf{G}^{p-1}_{1,2}(n) & \cdots & \mathbf{G}^{p-1}_{1,J^p}(n) \\
    \mathbf{G}^{p-1}_{2,1}(n) & \mathbf{G}^{p-1}_{2,2,:}(n) & \cdots & \mathbf{G}^{p-1}_{2,J^p}(n) \\
    \vdots & \vdots & \ddots & \vdots \\
    \mathbf{G}^{p-1}_{I^p,1}(n) & \mathbf{G}^{p-1}_{I^p,2}(n) & \cdots & \mathbf{G}^{p-1}_{I^p,J^p}(n)
\end{bmatrix},
\end{align}
where $I^p=I^{p-1}/l^p_i$ and $J^p=J^{p-1}/l^p_j$.
For instance, letting the side lengths be 4.
If the image at Stage $p-1$ is with size $64\times64$, the size becomes  $16\times16$ at Stage $p$.

\textbf{Step 2. Principal component analysis to local cuboids and feature reduction.}
The principal components are used as the transform kernels.
Principal component analysis (PCA) is a statistical procedure that uses an orthogonal transformation to convert a set of possibly correlated observations into linearly uncorrelated sets.
When data vectors are projected onto the principal components, the PCA coefficients have larger variances which can help in separating data into various classes.

First, we reshape the local cuboids $\mathbf{G}^{p-1}_{i,j}(n)$ to vectors $\mathbf{f}^p_{i,j}(n)\in R^{V^p}$, where $V^p$ is equal to the volume of the local cuboid defined as
\begin{align}
V^p=l^p_i\times l^p_j\times l^p_k,
\end{align}
and the reshape operation is defined as
\begin{align}\label{flatten}
\mathbf{f}^p_{i,j}(n)=\mathrm{reshape}(\mathbf{G}^{p-1}_{i,j}(n),V^p,1)\in R^{V^p}.
\end{align}
Second, the principal components of the vectors $\mathbf{f}^p_{i,j}(n)$ are calculated.
The principal components are the eigenvectors of the covariance matrix of the data vectors.
Let the mean for all data vectors be $\bar{\mathbf{f}}^p_{i,j}$ given by
\begin{align}
\bar{\mathbf{f}}^p_{i,j}=\frac{1}{N}\sum^N_{n=1}\mathbf{f}^p_{i,j}(n).
\end{align}
The covariance matrix  $\mathbf{R}^p_{i,j}$ is then calculated using
\begin{align}
\mathbf{R}^p_{i,j}=\frac{1}{N}\sum^N_{n=1}(\mathbf{f}^p_{i,j}(n)-\bar{\mathbf{f}}^p_{i,j})(\mathbf{f}^p_{i,j}(n)-\bar{\mathbf{f}}^p_{i,j})^T.
\end{align}
Then we find the $V^p$ eigenvectors $\mathbf{A}^p_{i,j}$ of $\mathbf{R}^p_{i,j}$, where each column of $\mathbf{A}^p_{i,j}$, denoted by ${[\mathbf{A}^p_{i,j}]}_v$, is the $v$th eigenvector of $\mathbf{R}^p_{i,j}$, and $v=1,2,\cdots,V^p$.
Also, let the corresponding eigenvalues of $\mathbf{R}^p_{i,j}$ be $\mathbf{e}^p_{i,j}$.
Hence ${[\mathbf{A}^p_{i,j}]}_v$ are the principal components of $\mathbf{f}^p_{i,j}(n)$.
Third,  the computational complexity of eigenvalues and eigenvectors is $\mathrm{O}(n^3)$, which is the main complexity in implementation.
Hence we propose to reduce the PCA coefficients according to the PCA property to reduce the complexity for later stages. Here, we select the  $K^p$ ($K^p \le V^p$) eigenvectors (principal components) corresponding to the $K^p$ largest eigenvalues,  which also have the largest variances and are considered to have significant discriminative ability.
Since the eigenvalues are equal to the variances that a vector projected onto its eigenvectors, this operation preserves the principal components corresponding to the largest variances.
More specifically,  one can find the set of indices ${\mathcal
T}$ corresponding to the $K^p$ largest eigenvalues defined as
\begin{align}\label{eigenvalue}
{\mathcal T}=\{t\,|\,[\mathbf{e}^p_{i,j}]_t\,\mathrm{are}\,\mathrm{the}\,K^p\,\mathrm{largest}\,\mathrm{eigenvalues}\}\in R^{K^p}.
\end{align}
Once the set ${\mathcal T}$ is found, the principal components can be reduced accordingly.
Then the transform kernels $\mathbf{B}^p_{i,j}$ after principal component reduction can be obtained by
\begin{align}\label{pcreduce}
\mathbf{B}^p_{i,j}=[\mathbf{A}^p_{i,j}]_{\mathcal T}\in R^{V^{p}\times K^{p}}.
\end{align}
Forth, we project the vectors $\mathbf{f}^p_{i,j}(n)$ onto the transform kernels (reduced principal components) and obtain the PCA coefficients $\mathbf{g}^p_{i,j}(n)$ defined as
\begin{align}\label{project}
\mathbf{g}^p_{i,j}(n)={\mathbf{B}^{p\;T}_{i,j}}\mathbf{f}^p_{i,j}(n)\in R^{K^p}.
\end{align}

\textbf{Step 3. Reshape PCA coefficients and form another global cuboid $\mathbf{G}^{p}(n)$.}
We then reshape all the PCA coefficients obtained in Step~2 to place the coefficient vectors in the spectral direction using the following procedure:
\begin{align}\label{reshape}
g^p_{i,j,:}(n)=\mathrm{reshape}(\mathbf{g}^p_{i,j}(n),1,1,K^p).
\end{align}
Combining all $g^p_{i,j,:}(n)$, where $i=1,2,\cdots,I^p$ and $j=1,2,\cdots,J^p$, global cuboid $\mathbf{G}^{p}(n)$ for next stage can be formed given by
\begin{align}\label{combine}
\mathbf{G}^{p}(n)=
\begin{bmatrix}
    g^p_{1,1,:}(n) & g^p_{1,2,:}(n) & \cdots & g^p_{1,J^p,:}(n) \\
    g^p_{2,1,:}(n) & g^p_{2,2,:}(n) & \cdots & g^p_{2,J^p,:}(n) \\
    \vdots & \vdots & \ddots & \vdots \\
    g^p_{I^p,1,:}(n) & g^p_{I^p,2,:}(n) & \cdots & g^p_{I^p,J^p,:}(n)
\end{bmatrix},
\end{align}
where $\mathbf{G}^{p}(n)\in R^{I^p\times J^p\times K^p}$.
The vertical and horizontal direction dimension conversion of global cuboids from Stage $p-1$ to Stage $p$ can be written as
\begin{align}
&I^p=I^{p-1}/l^p_i,\\
&J^p=J^{p-1}/l^p_j.
\end{align}

\textbf{Stop criterion.} The whole process is stopped when the final stage, Stage $P$, is approached,
and the global cuboids $\mathbf{G}^{P}(n)\in R^{1\times 1\times K^P}$ is obtained.

\begin{figure}
    \centering
    \includegraphics[width=0.35\textwidth]{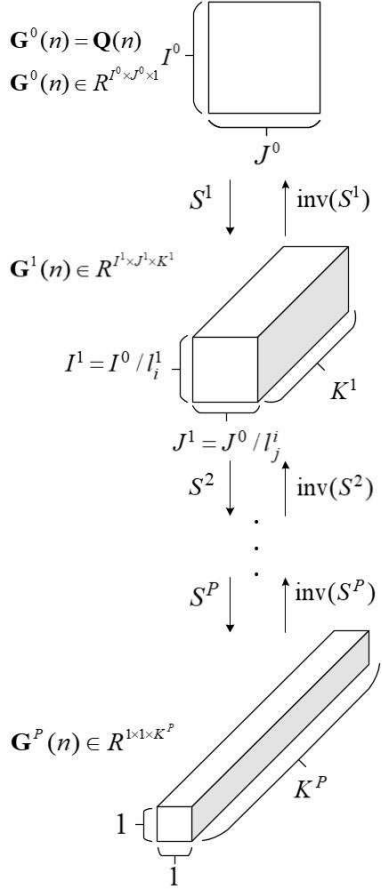}
    \caption{The dimension conversion of the global cuboid through stages. $S^p$ and $\
    \mathrm{inv}(S^p)$ are the forward and inverse of the proposed scheme between stage $p$ and $(p-1)$, respectively.}
	\label{dataflow}
\end{figure}

At the final stage, one can obtain the final PCA coefficients given by
\begin{align}
\mathbf{G}^{P}(n)=g^{P}_{1,1,:}(n)\in R^{1\times 1\times K^P},
\end{align}
where $K^P$ is normally set to a number that can avoid overfitting or underfitting in classification experiments so as to achieve high accuracy.
Also, since we reduce the PCA coefficients after each stage, $K^P$ is much smaller than the number of the original image pixels $(I\times J)$.
As a result, there is no growth but significant reduction in space complexity throughout the whole process.
Note that the space complexity doubles at each stage for the system in \cite{saak}.
Then the global cuboid at the final stage can be reshaped to obtain the feature vector $\mathbf{x}(n)$:
\begin{align}
\mathbf{x}(n)=\mathrm{reshape}(\mathbf{G}^{P}(n),K^P,1),
\end{align}
where  $\mathbf{x}(n)$ is called the feature vector, which is obtained from the PCA coefficients at the final stage, and is  used to determine the class of the input images later in classification component.

\subsection{Gaussian Approximation of Features}\label{gaussianapprox}

\begin{figure*}
    \centering
    \includegraphics[width=0.80\textwidth]{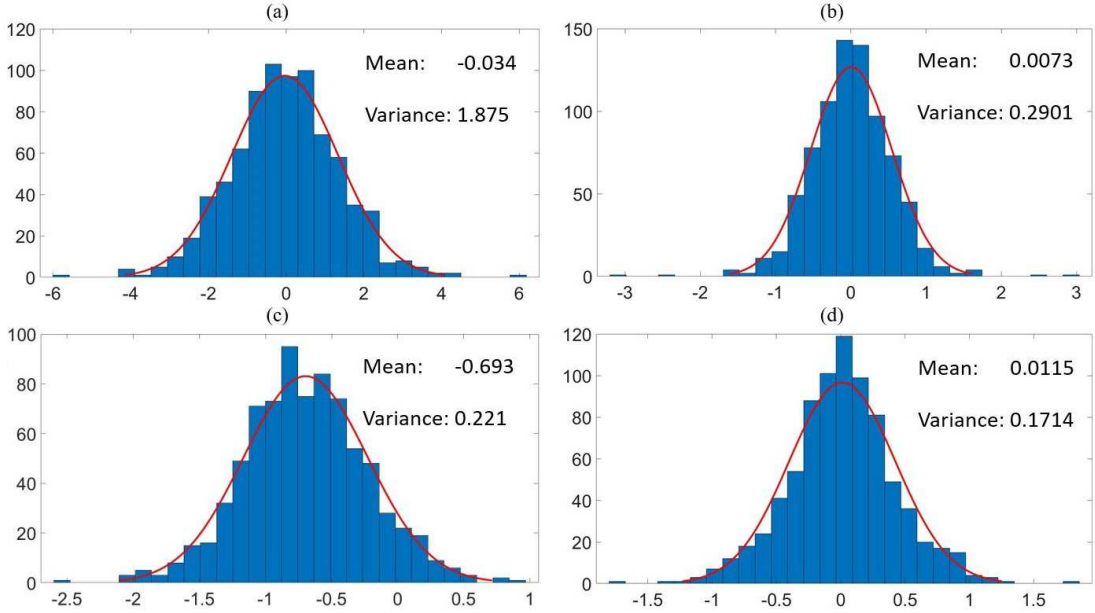}
    \caption{Gaussian approximation of the extracted features in the proposed system. (a) Feature No. 12, (b) Feature No. 55, (c) Feature No. 72, (d) Feature No. 90 from inputs  which belong to class 5.}
    \label{htotal}
\end{figure*}

\begin{figure}
    \centering
    \includegraphics[width=0.5\textwidth]{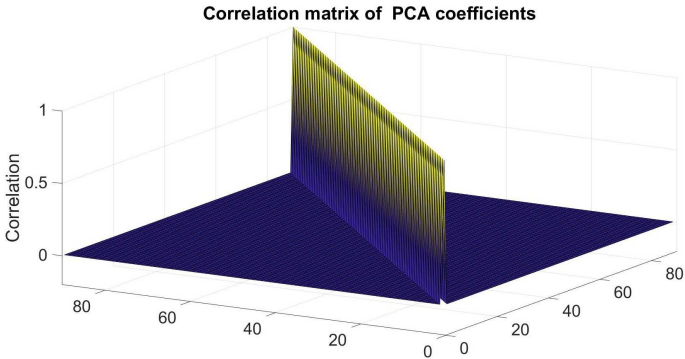}
    \caption{The correlation matrix of the extracted features, which is nearly diagonal with diagonal elements being 1 and the off-diagonal elements being smaller than 1.0e-05.}
    \label{corrmatrix}
\end{figure}

As mentioned in Step~2 of the previous subsection, the PCA is a procedure that can convert a set of possibly correlated coefficients into uncorrelated ones.
Hence the coefficients at the final stage should be almost uncorrelated.
To see this, the $(i,j)-th$ element of the correlation coefficient matrix $\rho_{\mathbf{x},\mathbf{x}}$ of the feature vector  is defined as
\begin{align}\label{rho}
\notag
&{[\rho_{\mathbf{x},\mathbf{x}}]}_{i,j}\\
&=\frac{\sum^N_{n=1}([\mathbf{x}(n)]_i-[\bar{\mathbf{x}}]_i)([\mathbf{x}(n)]_j-[\bar{\mathbf{x}}]_j)}{\sqrt{\sum^N_{n=1}{([\mathbf{x}(n)]_i-[\bar{\mathbf{x}}]_i)}^2\sum^N_{n=1}{([\mathbf{x}(n)]_j-[\bar{\mathbf{x}}]_j)}^2}},
\end{align}
where $\bar{\mathbf{x}}$ is the mean of $\mathbf{x}(n)$ defined as
\begin{align}\label{gsmean}
\bar{\mathbf{x}}=\frac{1}{N}\sum^N_{n=1}\mathbf{x}(n).
\end{align}
For example, let $\mathbf{x}$  be the feature vector that will be detailed in Experiment 1 in Sec.~\ref{secers}, where the dimension is 90.
Then, the obtained correlation coefficient matrix is shown in Fig.~\ref{corrmatrix}.
From the figure, we see that the PCA coefficients at the final stage are almost uncorrelated.
In addition to the uncorrelated relationship among features, we also find that the statistics of individual features can be approximated by Gaussian distribution.
The histograms of some randomly picked samples are shown in Fig.~\ref{htotal}, in which Gaussian approximation well matches the histograms.
In addition to this dataset, we have also verified various datasets and observed similar Gaussian approximation results of the proposed system.

Knowing that the features are nearly uncorrelated Gaussian distribution, we can treat the features as the Gaussian mixture model (GMM).
The GMM phenomenon was also reported in \cite{saak}.
Approximating the features using the GMM is useful and motivates us to use the linear discriminant classifier, a MAP-based detector, that will be introduced in Sec.~$\mathrm{V}$.

\subsection{Inverse Direction}\label{secinverse}
Since the proposed scheme is a linear transformation, both forward and inverse transformations can be conducted.
The inverse transformation is to reconstruct the feature vector $\mathbf{x}(n)$ at the final stage to the preprocessed image $\mathbf{Q}(n)$.
We elaborate on  the inverse transformation in the following steps:
Now the stage index $p$ is with decreasing order, {\em i.e.,} $p=P, P-1, \cdots, 1$.

\textbf{Step 1. Reshape the PCA coefficients back to vectors $\mathbf{g}^{p}_{i,j}(n)$.}
This is a simple inverse procedure of (\ref{reshape}) given by
\begin{align}
\mathbf{g}^{p}_{i,j}(n)=\mathrm{reshape}(g^p_{i,j,:}(n),K^p,1).
\end{align}

\textbf{Step 2. Project the the vectors onto the inverse transform kernel.}
Referring to (\ref{project}),
the inverse kernel of ${\mathbf{B}^{p\;T}_{i,j}}$ is simply its pseudo-inverse  matrix.
Hence the backward result of this step can be easily obtained by
\begin{align}
\mathbf{f}^{p}_{i,j}(n)=({\mathbf{B}^{p\;T}_{i,j}})^\mathrm{pinv}\mathbf{g}^{p}_{i,j}(n),
\end{align}
where $({\mathbf{B}^p_{i,j}}^T)^\mathrm{pinv}$ is the pseudo-inverse matrix of ${\mathbf{B}^p_{i,j}}^T$.

\textbf{Step 3. Reshape $\mathbf{f}^{p}_{i,j}(n)$ back to the local cuboids.}
Referring to (\ref{flatten}), this step is given by
\begin{align}
\mathbf{G}^{p-1}_{i,j}(n)=\mathrm{reshape}(\mathbf{f}^{p}_{i,j}(n),l^{p}_i,l^{p}_j,l^{p}_k).
\end{align}

\textbf{Step 4. Form global cuboid and take PCA coefficients.}
The method to collect the local cuboids to form global cuboid is the inverse of (\ref{decomposition}), and the method for taking the PCA coefficients for the previous stage is the inverse of (\ref{combine}).

\begin{figure}
	\centering
	\includegraphics[width=0.45\textwidth]{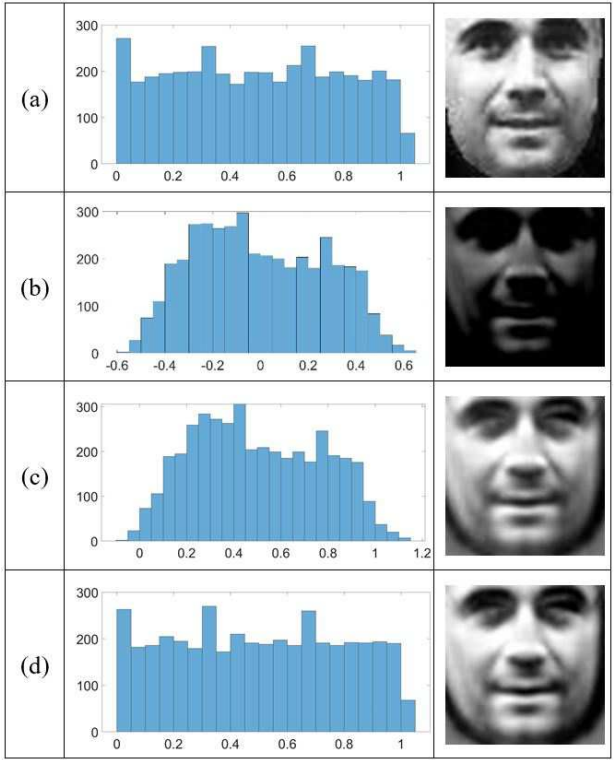}
	\caption{(a): Original image and its histogram. (b): Recovered image without any image processing. (c): Recovered image after adding brightness. (d): Recovered image after adding brightness and histogram equalization.}
	\label{ip}
\end{figure}

Conduct Steps 1-4 for all $P$ stages, the initial global cuboid $\hat{\mathbf{G}}^{0}(n)$ can be reconstructed.
Due to the energy loss from the eliminated features,  image processing is needed for reducing the reconstruction loss.
First, we compensate the energy loss using a brightness gap $h(n)$ defined as
\begin{align}\label{brightness}
h(n)=\frac{1}{I\cdot J}\sum^I_{i=1}\sum^J_{j=1}([\mathbf{Q}(n)]_{i,j}-[\hat{\mathbf{G}}^{0}(n)]_{i,j}).
\end{align}
Second, we apply histogram equalization to enhance the contrast and fix the range span as well.
The histogram equalization was widely used, see {\em e.g.,} \cite{histomri} and \cite{anovel}.
Here, we use the function $\mathrm{histeq}$ provided by the Matlab.
Finally, the recovered image can be obtained with a slight loss $\hat{\mathbf{Q}}(n)$ defined as
\begin{align}\label{riip}
\hat{\mathbf{Q}}(n)=\mathrm{histeq}(\hat{\mathbf{G}}^{0}(n)+h(n)).
\end{align}
In Fig. \ref{ip}, we provide an example using an image from the testing dataset that will be introduced in Sec.~\ref{secers} later.
The original features of the image in the red layer is 4096, and we recover the image using only 90 features.
In row (a), the red layer of the original image and its histogram are shown.
In row (b), we show the recovered image without any image processing $\hat{\mathbf{G}}^{0}(n)$ and we can see the image is darker than the original one and the contrast is poor, though one can still recognize the image.
In row (c), the brightness has been compensated  by adding $h(n)$ defined in (\ref{brightness}).
Hence the mean of the histogram shifts.
In row (d), we show the recovered image $\hat{\mathbf{Q}}(n)$ using (\ref{riip}). One can see that no matter from the image or its histogram, they are the closest to those in row (a).
Thus the image processing in (\ref{brightness}) and (\ref{riip}) really helps when recovering images from reduced few features.

When $K^P<(I\times J)$, reconstructing images from the reduced features can be regarded as a lossy data decompression.
Such data compression and decompression are meaningful for data classification.
More specifically, if only $K^P$ most significant features are needed to achieve a target classification performance, one can simply store the data set with $K^P$ features instead of $(I\times J)$ features because both have comparable classification performance.
Namely, we only store the top $K^P$ significant features $\mathbf{x}(n)\in R^{K^P}$, instead of the original data $\mathbf{Q}(n)\in R^{I\times J}$.
Such a concept may be used to reduce the data size and computational complexity in classification algorithms.
The compression ratio $r$ with feature reduction can then be written as
\begin{align}
r=\frac{I\times J}{K^P}.
\end{align}
The examples in Fig. \ref{ip} (a) and (d) thus have a compression ratio of $4096:90=45.51:1$.

\section{Linear Discriminant Classifier}

After the feature extraction, the features can be fed into the classifier, which is introduced here.
As discussed in Sec.~\ref{gaussianapprox}, the features can be approximated to be uncorrelated Gaussian random variables.
Hence classical theory in estimation and detection can be used and the resulting performances have theoretical explanations.
Now we have the reduced feature vector $\mathbf{x}(n)$ and would like to determine which class it belongs to.
To solve such a problem, classical maximum a posteriori (MAP) estimator provides a good reference.
In the area of machine learning and pattern recognition, this concept is usually called linear discriminant analysis (LDA), which finds a linear combination of features that can separate two or more classes.
Here we combine the two concepts,  LDA and MAP estimator to design the linear discriminant classifier introduced as follows:

First,  let $y(n)$ be the output of the classifier, which is the class that $\mathbf{x}(n)$ belongs to, and $C_m$ be Class $m$.
The MAP estimator is given by
\begin{align}\label{original classifier}
y(n)&=\arg\max_{m=1\cdots M}P(C_m|\mathbf{x}(n)),
\end{align}
where $M$ is the number of classes.
The posterior probability is defined as
\begin{align}\label{posteriorprobability}
P(C_m|\mathbf{x}(n))=\frac{P(\mathbf{x}(n)|C_m)P(C_m)}{P(\mathbf{x}(n))},
\end{align}
where $P(\mathbf{x}(n))$ is written as
\begin{align}\label{px}
P(\mathbf{x}(n))=\sum_{m=1}^{M}P(\mathbf{x}(n)|C_m)P(C_m).
\end{align}
Since $P(\mathbf{x}(n))$ is a constant for all classes, one can ignore the denominator in (\ref{posteriorprobability}).
As for the numerator in (\ref{posteriorprobability}), since we consider the input features are multivariate Gaussian distributed, the priori probability can be written as
\begin{align}\label{pc}
P(C_m)=\frac{n_m}{N},
\end{align}
where $n_m$ is the number of training images which are with class $m$, and $N=\sum^M_{m=1}n_m$ is the number of all training images.
The mean of $\mathbf{x}_m(n)$ is defined as
\begin{align}
\mathbf{\mu}_m=\frac{1}{n_m}\sum^{n_m}_{n=1}\mathbf{x}_m(n),
\end{align}
and the covariance matrix $\mathbf{\Sigma}_m$ is defined as
\begin{align}\label{covariance}
\mathbf{\Sigma}_m=\frac{1}{n_m}\sum_{n=1}^{n_m}(\mathbf{x}_m(n)-\boldsymbol{\mu}_m)^T(\mathbf{x}_m(n)-\boldsymbol{\mu}_m),
\end{align}
where $\mathbf{x}_m(n)$, $n=1,2,\cdots,n_m$, are the feature vectors belong to class $m$.

In the linear discriminant classifier, we need a common covariance matrix so as to keep the likelihood function linear.
In this work, we propose to use the pooled covariance matrix $\mathbf{\Sigma}$ as the common matrix.
The pooled covariance matrix is defined as
\begin{align}\label{pooledcovariance}
\mathbf{\Sigma}=\sum_{m=1}^{M}\frac{n_m}{N}\mathbf{\Sigma}_m.
\end{align}
The likelihood function $P(\mathbf{x}(n)|C_m)$ for multivariate Gaussian random variables is then given by
\begin{align}\label{pxc}
&P(\mathbf{x}(n)|C_m)=\\
&\frac{1}{\sqrt{(2\pi)^{K^P}|\mathbf{\Sigma}|}}\exp\left(-\frac{1}{2}(\mathbf{x}(n)-\boldsymbol{\mu}_m)^T\mathbf{\Sigma}^{-1}(\mathbf{x}(n)-\boldsymbol{\mu}_m)\right),
\end{align}
where recall that $K^P$ is the number of reduced  features.

To simplify the calculation,
referring to (\ref{posteriorprobability}),
we take the natural logarithm, and ignore the denominator of the posterior probability function $P(C_m|\mathbf{x}(n))$, {\em i.e.,}
\begin{align}\label{y simplify}
y(n)&=\arg\max_{m=1\cdots M}\ln{P(\mathbf{x}(n)|C_m)P(C_m)},
\end{align}
The term $\ln P(\mathbf{x}(n)|C_m)P(C_m)$ can be rewritten as
\begin{align}\label{derivation of prob}
&\notag\ln P(\mathbf{x}(n)|C_m)P(C_m)\\
&=\ln(\frac{1}{\sqrt{(2\pi)^{K^P}|\mathbf{\Sigma}|}})\notag\\
&-\frac{1}{2}(\mathbf{x}(n)^T\mathbf{\Sigma}^{-1}\mathbf{x}(n)-2\boldsymbol{\mu}_m^T\mathbf{\Sigma}^{-1}\mathbf{x}(n)+\boldsymbol{\mu}_m^T\mathbf{\Sigma}^{-1}\boldsymbol{\mu}_m)\notag\\
&+\ln(\frac{n_m}{N}).
\end{align}
From (\ref{pooledcovariance}), since the covariance is common, $\ln(\frac{1}{\sqrt{(2\pi)^{K^P}|\mathbf{\Sigma}|}})$ and $-\frac{1}{2}\mathbf{x}(n)^\mathrm{T}\mathbf{\Sigma}^{-1}\mathbf{x}(n)$ can be ignored.
Then the remaining term can be written in a linear form consisting of a weight $\mathbf{w}_m$ and a bias term $b_m$ for an individual class given by
\begin{align}\label{linear form}
\boldsymbol{\mu}_m^T\mathbf{\Sigma}^{-1}\mathbf{x}(n)-\frac{1}{2}\boldsymbol{\mu}_m^T\mathbf{\Sigma}^{-1}\boldsymbol{\mu}_m+\ln({\frac{n_m}{N}})=\mathbf{w}_m\mathbf{x}(n)+b_m ,
\end{align}
where the term $\mathbf{w}_m=\boldsymbol{\mu}_m^T\mathbf{\Sigma}^{-1}$ is the weight and the term $b_m=-\frac{1}{2}\boldsymbol{\mu}_m^T\mathbf{\Sigma}^{-1}\boldsymbol{\mu}_m+\ln({\frac{n_m}{N}})$ is the bias.
Finally, the classifier can be written in a linear equation form as follows:
\begin{align}\label{y linear form}
y(n)=\arg\max_{m=1\cdots M}\mathbf{w}_m\mathbf{x}(n)+b_m\;.
\end{align}

We summarize the proposed scheme for training and testing including the linear transformation of feature extraction, feature reduction and linear discriminant classifier in respectively in Algorithms~\ref{Algo:inner} and \ref{Algo:inner2}.

\begin{algorithm}\label{Algo:inner}
\caption{Training Process}
\begin{algorithmic}[1]
\REQUIRE $N$ preprocessed training images $\mathbf{Q}(n)\in R^{I\times J}$. Local cuboid spatial sizes $l^p_i\times l^p_j$,
and global cuboid size at spectral direction $K^p$
at stage $p$ for $p=1,2,\cdots,P$.
\ENSURE Transform kernels $\mathbf{B}^p_{i,j}$. Weight $\mathbf{w}_m$ and bias $b_m$. Remaining indices ${\mathcal T}$.
\STATE \textbf{Initialization:} $\mathbf{G}^{0}(n)=\mathbf{Q}(n)$. Let global cuboid size be $I^0\times J^0\times K^0$, where $I^0=I$ , $J^0=J$ and $K^0=1$.
\begin{center}- - - - - - - - - - - \textbf{Feature Extraction} - - - - - - - - - - -\end{center}
\FOR{$p=1:P$}
    \STATE Form local cuboids $\mathbf{G}^{p-1}_{i,j}(n)$ using (\ref{decomposition}) and reshape them to vectors $\mathbf{f}^p_{i,j}(n)$ using (\ref{flatten}).
    \STATE $I^p=I^{p-1}/l^p_i$ and $J^p=J^{p-1}/l^p_j$
    \FOR{$i=1:I^p$}
        \FOR{$j=1:J^p$}
        %
        %
        \STATE Find $K^p$ transform kernels (reduced PCs) $\mathbf{B}^p_{i,j}$ using eigenvector-based method in (\ref{eigenvalue}) and (\ref{pcreduce}).
        \STATE Project $\mathbf{f}^p_{i,j}(n)$ onto $\mathbf{B}^p_{i,j}$ to obtain PCA coefficients $\mathbf{g}^p_{i,j}(n)$ using (\ref{project}).
        \STATE Reshape $\mathbf{g}^p_{i,j}(n)$ to $g^p_{i,j,:}(n)$ to form global cuboid $\mathbf{G}^{p}(n)$ using (\ref{combine}).
        \ENDFOR
    \ENDFOR
\ENDFOR
\STATE Obtain final PCA coefficients $g^P_{i,j,:}(n)$ at the final stage and reshape them to feature vector $\mathbf{x}(n)$.
\STATE For colored images, repeat 1-15 for each primary color layer, and cascade all features to a vector $\mathbf{x}(n)$.
\begin{center}- - - - - - - \textbf{Linear Discriminant Classifier} - - - - - - -\end{center}
\STATE Feed $\mathbf{x}(n)$ and labels into the Linear Discriminant Classifier, and obtain weight $\mathbf{w}_m$ and bias $b_m$.
\end{algorithmic}
\end{algorithm}

\begin{algorithm}\label{Algo:inner2}
\caption{Testing Process}
\begin{algorithmic}[1]
\REQUIRE The $N$ preprocessed testing image(s) $\mathbf{Q}(n)\in R^{I\times I}$. Local cuboid spatial sizes $l^p_i\times l^p_j$ same as Algo. \ref{Algo:inner}. Transform kernels $\mathbf{B}^p_{i,j}$. Weight $\mathbf{w}_m$ and bias $b_m$. Remaining indices ${\mathcal T}$.
\ENSURE The classification accuracy.
\STATE \textbf{Initialization:} $\mathbf{G}^{0}(n)=\mathbf{Q}(n)$. Let global cuboid size be $I^0\times J^0\times K^0$, where $I^0=I$ , $J^0=J$ and $K^0=1$.
\begin{center}- - - - - - - - - - - \textbf{Feature Extraction} - - - - - - - - - - -\end{center}
\FOR{$p=1:P$}
    \STATE Repeat Steps 3-4 in Algo. \ref{Algo:inner}.
    \FOR{$i=1:I^p$}
        \FOR{$j=1:J^p$}
        \STATE Read transform kernels $\mathbf{B}^p_{i,j}$.
        \STATE Repeat Steps 8-9 in Algo. \ref{Algo:inner}.
        \ENDFOR
    \ENDFOR
\ENDFOR
\STATE Repeat Step 13-14 in Algo. \ref{Algo:inner}.
\begin{center}- - - - - - - \textbf{Linear Discriminant Classifier} - - - - - - -\end{center}
\STATE Use the classifier in
(\ref{y linear form})
to estimate which class the $n$th image belongs to.
\STATE Use label(s) to check the classification accuracy.
\end{algorithmic}
\end{algorithm}

\section{Experimental Results}\label{secers}
In this section, the experiment results are provided to show the performance of the proposed classification system.

\textbf{Dataset and Hardware.} Labeled Faces in the Wild (LFW) is a well known academic test set for face verification
\cite{facenet}.
There are two datasets used in the experiments.
For the first data set, 158 classes are selected from the whole dataset which has more than 6 images for training and 2 images to testing per class.
For the second data set, 19 classes are selected out of the 158 classes in the first data set,
which has more than 30 images for training and 10 images for testing.
Table~\ref{dts} details the size of the two datasets.
Intel(R) Core(TM) i7-8700CPU and 16GB RAM is used to conduct the experiments.

\begin{table}[!htb]
\centering
\caption{Dataset Details}
\begin{tabular}{|c|c|c|c|c|}
\hline
$\#$ & Training data & Testing data & Total & Total \\
classes & per class & per class & training data& testing data\\ \hline
$19$ & $\geq 30$ & $\geq 10$ & $1402$ & $460$\\ \hline
$158$ & $\geq 6$ & $\geq 2$ & $3296$ & $1014$\\ \hline
\end{tabular}
\label{dts}
\end{table}

\begin{figure}
    \centering
    \includegraphics[width=0.5\textwidth]{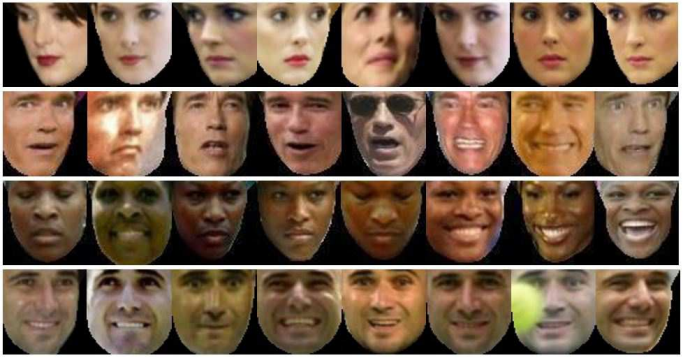}
    \caption{An overview of images after face detection and resizing.}
    \label{dataset}
\end{figure}

\textbf{Image Preprocess.}
Firstly, the face is detected and the background of the image is blacked via an open-source python file
\cite{shape}.
Then  all the images are resized to $64\times 64$.
An overview of the images after the face detection operation and resizing is shown in Fig.~\ref{dataset}.
Secondly,  the training images are augmented.
The reason is that in the first dataset, there are some classes that have only 6 images for training, which is insufficient to train a good model.
Also, some faces do not face the same directions as shown in Fig.~\ref{dataset}.
Therefore, we augment the training images by flipping the images horizontally and make the number of images doubles.
Thirdly, we separate the three primary colors of images into R, G, B 3 layers and then equalize the 3 layers of histogram individually.
This step enhances the contrast of images and enable the separation of the three primary color layers.

After preprocessing, the training data are passed through the proposed system, and Algorithm~\ref{Algo:inner} is used to process the data.

\textbf{Experiment 1: Testing accuracy as functions of various numbers of features.}
As discussed in Sec.~\ref{secinverse},
the number of features id reduced from $(I\times J)$ to $K^P$ to avoid overfitting in training classifier, where $I\times J=4096$ for one primary color layer in the experiments.
For designing a face recognition model, the main goal is to achieve high testing accuracy, and thus $K^P$ is decided mainly according to the testing accuracy.
Let the local cuboid sizes be set as $(l^1_i,l^1_j,l^1_k)=(8,8,1)$, $(l^2_i,l^2_j,l^2_k)=(4,4,16)$, $(l^3_i,l^3_j,l^3_k)=(2,2,64)$, which is a 3-stage scheme.
Table~\ref{pldc} shows the testing accuracy corresponding to 8 different numbers of features.
Several observations are summarized: First, more features do not necessarily lead to better testing accuracy.
This is because more features may result in overfitting problem and hence feature reduction is needed.
Second, for both datasets, the best testing accuracy occurs when $K^P=90$ for one primary color layer and  a total of 270 features for three layers.
Consequently, the compression ratio is  $1:45.5$, which is a significant reduction of data size.
\begin{table}[!htb]
\centering
\caption{Testing accuracy as functions of various numbers of features}
\begin{tabular}{|a|c|c|a|c|c|}
\hline
$\#$ features & \multicolumn{2}{|c|}{$\#$ classes} & $\#$ features & \multicolumn{2}{|c|}{$\#$ classes}\\ \hline
& 19 & 158 & & 19 & 158\\ \hline
12288& 12.61\%& 0.10\%& 300& 97.17\%& 83.53\%\\ \hline
9000& 5.43\%& 0.10\%& \textbf{270}& $\mathbf{97.61\%}$& $\mathbf{84.91\%}$\\ \hline
6000& 7.39\%& 23.08\%& 240& 97.17\%& 84.12\%\\ \hline
3000& 6.09\%& 61.24\%& 90& 93.91\%& 78.60\%\\ \hline
900& 96.09\%& 81.66\%& 45& 87.61\%& 63.51\%\\ \hline
\end{tabular}
\label{pldc}
\end{table}

\textbf{Experiment 2: Testing accuracy and computations as functions of various local cuboid sizes.}
In this experiment, three sets of variable local cuboid sizes are used to figure out efficient local cuboid sizes.
Efficiency considers the computational speed and testing accuracy.
The computational speed depends on the number of stages.
When the local cuboid sizes are large, the proposed multi-stage scheme has  fewer stages and hence fewer computations, but the feature extraction might lose the precision of extracting the most significant features, and it leads to bad accuracy in the end.
Thus, a good local cuboid size should be determined for achieving high testing accuracy with low computations.
We consider the following three settings of local cuboid sizes:
\textbf{Setting~1:} There is one stage with $(l^1_i,l^1_j,l^1_k)=(64,64,1)$.
\textbf{Setting~2:} There are 2 stages with $(l^1_i,l^1_j,l^1_k)=(4,4,1)$, $(l^2_i,l^2_j,l^2_k)=(16,16,1)$.
\textbf{Setting~3:} There are 3 stages with $(l^1_i,l^1_j,l^1_k)=(8,8,1)$, $(l^2_i,l^2_j,l^2_k)=(4,4,16)$, $(l^3_i,l^3_j,l^3_k)=(2,2,64)$.
\textbf{Setting~4:} There are 6 stages with $(l^1_i,l^1_j,l^1_k)=(2,2,1)$, $(l^2_i,l^2_j,l^2_k)=(2,2,1)$, $(l^3_i,l^3_j,l^3_k)=(2,2,1)$, $(l^4_i,l^4_j,l^4_k)=(2,2,2)$, $(l^5_i,l^5_j,l^5_k)=(2,2,8)$, $(l^6_i,l^6_j,l^6_k)=(2,2,32)$.
Here, recall that $l^p_k=K^{p-1}$ and can be set for individual stages.
Also we let $K^P=90$, the same setting as that in Experiment 1.
Thus we can obtain 270 features totally from three primary color layers.
The cuboid sizes  of individual Settings are listed in Table~\ref{gcs}.
The results are shown in Table~\ref{vls19}.
Observed from the table that Settings 3 achieves the best accuracy for both datasets.
Also, Setting 3 has a lower computational complexity than that in Settings 1 and 4.
Hence, Setting 3 is a good setting for the experiments.
Some interesting results are also addressed here.
First, Setting 1 has the largest computational complexity since this setting only has one stage and the data is not segmented into local cuboids.
Consequently the computations for eigenvectors and eigenvalues dominate the complexity.
Moreover, although Setting 2 can achieve low computational time, its testing accuracy is far worse than other settings, because the PCA coefficients are seriously eliminated after the first stage, and much information is lost due to this.
Hence, the cuboid sizes  should be properly determined, like what we have done for Setting 3.
\begin{table*}[!htb]
	\centering
	\caption{The cuboid sizes of individual Settings.}
	\begin{tabular}{|c|c|c|c|c|c|c|}
		\hline
		Stage& 1 & 2 & 3 & 4 & 5 & 6 \\ \hline
		Setting 1 & (1,1,90) & \multicolumn{5}{|c|}{}\\ \hline
		Setting 2 & (16,16,1) & (1,1,90) & \multicolumn{4}{|c|}{}\\ \hline
		Setting 3 & (8,8,16) & (2,2,64) & (1,1,90) & \multicolumn{3}{|c|}{}\\ \hline
		Setting 4 & (32,32,1) & (16,16,1) & (8,8,2) & (4,4,8) & (2,2,32) & (1,1,90)\\ \hline
	\end{tabular}
	\label{gcs}
\end{table*}

\begin{table*}[!htb]
\centering
\caption{Testing accuracy and computations as functions of various local cuboid sizes.}
\begin{tabular}{|c|c|c|c|c|c|c|}
\hline
 & \multicolumn{3}{|c|}{19-classes} & \multicolumn{3}{|c|}{158-classes} \\ \hline
Setting & Time(Train) & Time(Test) & Accuracy & Time(Train) & Time(Test) & Accuracy\\ \hline
1 & $9.3$s & $0.11$s & $97.61\%$ & $10.4$s & $0.24$s & $84.71\%$\\ \hline
2 & $0.7$s & $0.04$s & $92.39\%$ & $1.7$s & $0.13$s & $78.69\%$\\ \hline
\textbf{3} & $\mathbf{0.7}$s & $\mathbf{0.06}$s & $\mathbf{97.61\%}$ & $\mathbf{1.7}$s & $\mathbf{0.14}$s & $\mathbf{84.91\%}$\\ \hline
4 & $1.9$s & $0.2$s & $95.86\%$ & $4.1$s & $0.54$s & $83.33\%$\\ \hline
\end{tabular}
\label{vls19}
\end{table*}

\begin{figure}
    \centering
    \includegraphics[width=0.5\textwidth]{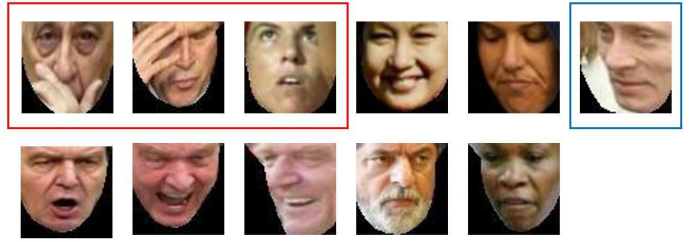}
    \caption{All the 11 incorrectly recognized images from the 19-classes dataset. There are some problematic images boxed in red and blue rectangles. If they are eliminated, the accuracy can reach $98.5\%$}
    \label{err}
\end{figure}

\begin{figure}
    \centering
    \includegraphics[width=0.5\textwidth]{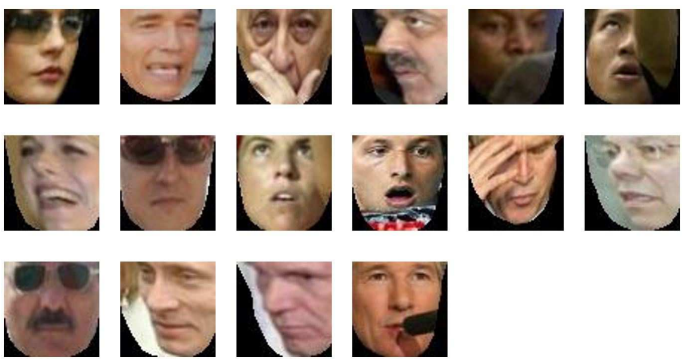}
    \caption{All the incorrectly recognized problematic images from the 158-classes dataset. If these unavoidable errors are eliminated, the accuracy can reach $86.3\%$.}
    \label{err158}
\end{figure}

\textbf{Experiment 3: Performance with multiple candidates.}
Following Experiments 1 and 2, we show  all the images that are incorrectly recognized from the 19-classes testing dataset in Fig.~\ref{err}.
We see that there are some problematic images which are marked with red and blue rectangles.
In the images with red rectangles, the faces are covered by hands; while in the image with a blue rectangle, the error actually came from image preprocessing (face detection operation).
When we eliminate those problematic images or select a better face detection operation, the testing accuracy can be improved from 97.6$\%$ to 98.5$\%$.
Similarly, in Fig.~\ref{err158}, we displayed the incorrectly recognized problematic images from the 158-classes testing dataset.
Some of them are seriously affected by sunglasses, hands, and even other's shoulder.
When we exclude these errors, the accuracy improves from 84.9$\%$ to 86.5$\%$.

\begin{figure}
    \centering
    \includegraphics[width=0.5\textwidth]{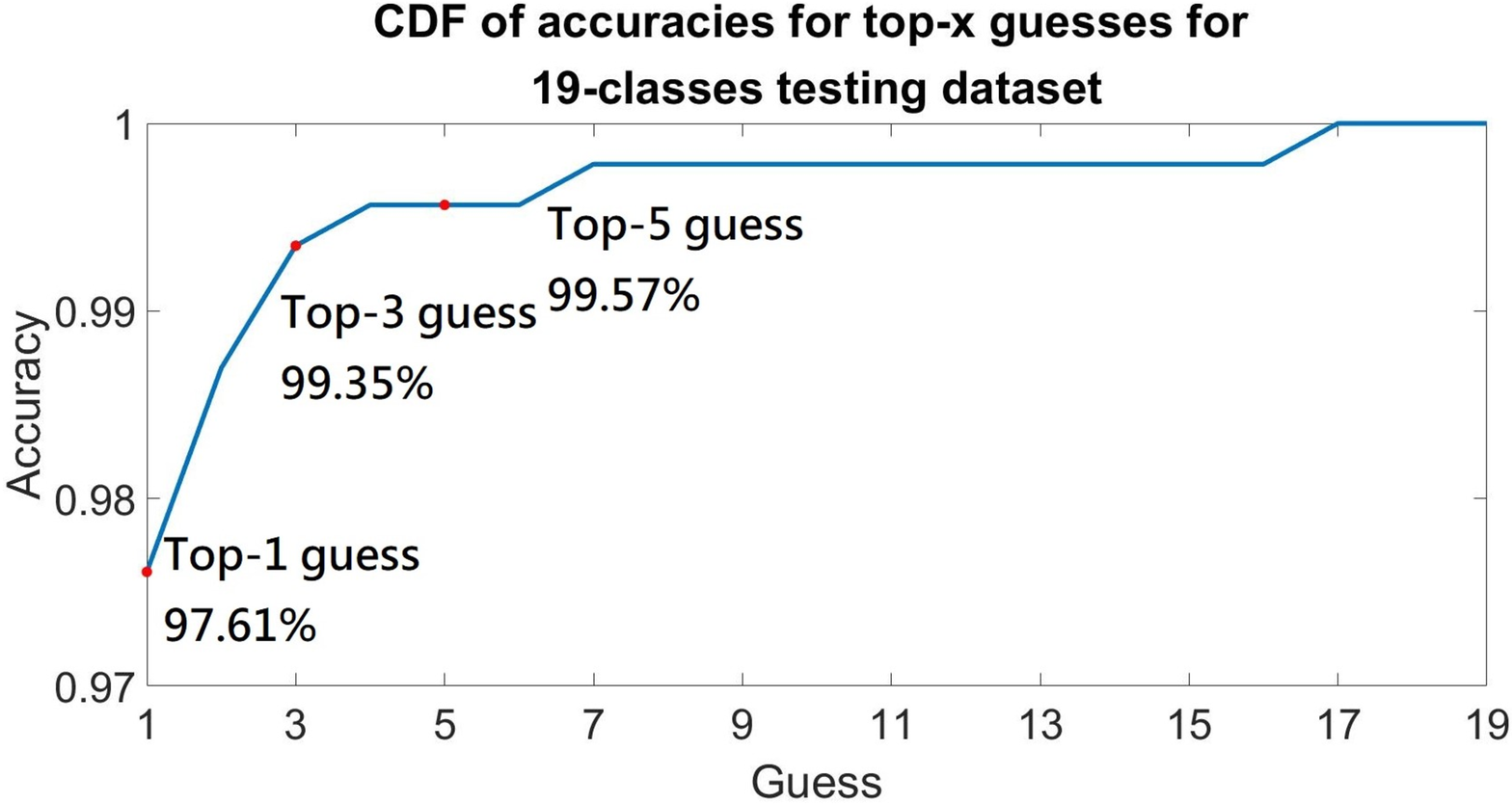}
    \includegraphics[width=0.5\textwidth]{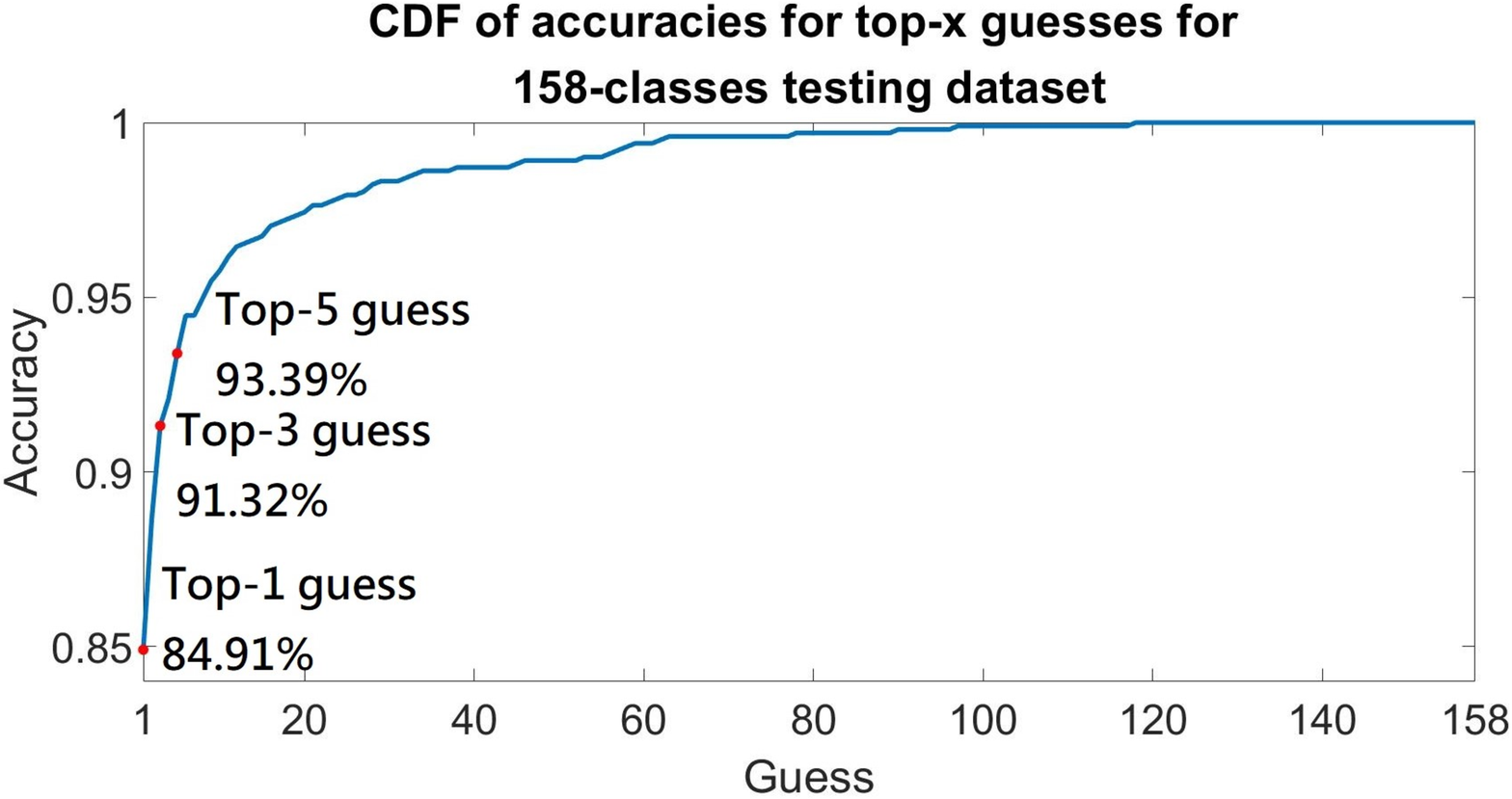}
    \caption{The CDF of top-X guesses for both testing datasets. The accuracies of both datasets reach above $90\%$ when having top-3 guesses, $99.35\%$ for the 19-class dataset and $91.32\%$ for the 158-class dataset.}
    \label{cdfs}
\end{figure}

Additionally, because the proposed scheme classifies the images using a MAP-like estimator, for each image the classifier outputs the individual probabilities for individual classes that this image belongs to.
Hence it is possible to determine several candidates with the highest probabilities that one image belongs to.
Refined and improved algorithms can be developed to determine the best decision from the candidates with the highest probabilities.
It is worth mentioning that some classifiers cut space into regions and do not have the cluster center in advance, such as SVM.
Such schemes need to pay more effort if more candidates are to be selected.
Here we show the testing accuracy of top-3 and top-5 candidates.
The accuracy for 158-class dataset reaches 91.32$\%$ when having top-3 guesses, and 93.39$\%$ when having top-5 guesses.
Moreover, the accuracy for 19-class dataset reaches 99.35$\%$ when having top-3 guesses, and 99.57$\%$ when having top-5 guesses.
The CDFs of top guesses are shown in Fig. \ref{cdfs}.

\textbf{Experiment 4: Image reconstruction using various numbers of features.} In this experiment, we reconstruct images from various numbers of features.
The reasons are two folded.
The first one is to see how many features are sufficient to reconstruct the original images from a point of view of data compression.
The second reason is to see what features and why they are eliminated in the original images.
In Experiments 1 and 2, we see using full features may lead to overfitting problem.
By reversing the images from the reduced features, the corresponding results can also be used to explain what may lead to overfitting problem and what are the important features of the original images that can be used to mostly distinguish images from different classes.
Here, we recover images from 12288 (full features), 9000, 6000, 3000, 270 features which respectively have 4096, 3000, 2000, 1000, 90 features for recovering one primary color layer.
To compare the recovery result, we use the percent deviation to calculate the loss defined as
\begin{align}
l=\frac{\sqrt{\sum_{i=1}^{I^0}\sum_{j=1}^{J^0}\sum_{k=1}^{K^0}\left([\hat{\mathbf{Q}}(n)]_{i,j,k}-[\mathbf{Q}(n)]_{i,j,k}\right)^2}}{I^0\times J^0\times K^0}.
\end{align}
Table~\ref{cr} shows the corresponding percent deviations and compression ratio.
It is observed that using more features does not necessarily lead to a small percent deviation due to overfitting issue.
In addition,  reconstructing images from 270 features has a very high compression ratio and a satisfactory low percent deviation.
Note that the deviation for 12288 is very small but nonzero due to the accumulated computational error of multiple stages in the platform.
\begin{table}[!htb]
\centering
\caption{Compression ratios}
\begin{tabular}{|c|c|c|}
\hline
Features & Percent deviation & Compression ratios\\ \hline
12288 & 1.7e-04\% & 1:1\\ \hline
9000 & 5.48\% & 1.37:1\\ \hline
6000 & 12.41\% & 2.05:1\\ \hline
3000 & 15.51\% & 4.10:1\\ \hline
\textbf{270} & \textbf{9.11\%} & \textbf{45.51:1}\\ \hline
240 & 9.31\% & 51.2:1\\ \hline
90 & 11.45\% & 136.5:1\\ \hline
\end{tabular}
\label{cr}
\end{table}

\begin{figure}
    \centering
    \includegraphics[width=0.48\textwidth]{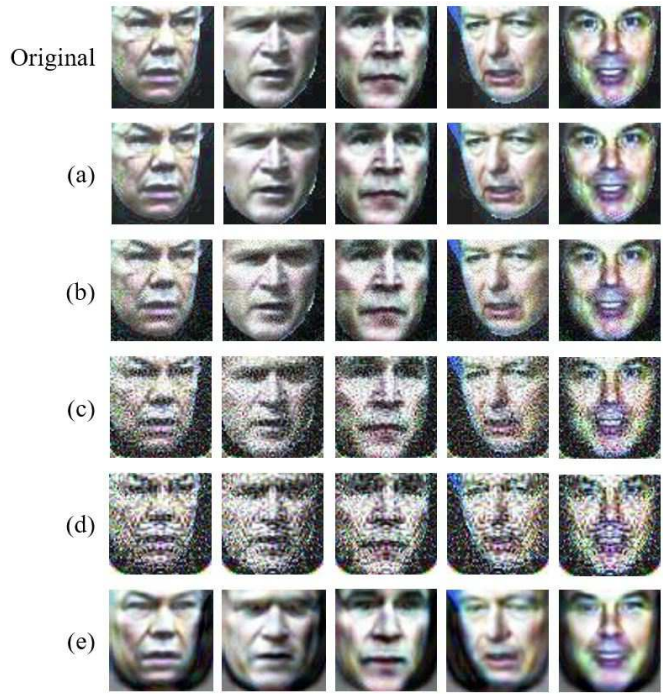}
    \caption{Some image recovery samples. (a) are the samples of recovery images recovered from 12288 features, the percent deviation is $1.7226e-04\%$. (b) to (e) are recovered from reduced features, and the numbers of features are 9000, 6000, 3000, 270, the percent deviations are $5.48\%, 12.41\%, 15.51\%, 9.11\%$.}
    \label{recovers}
\end{figure}

Let us see how the recovered images look like.
Fig. \ref{recovers}, shows some reconstructed sample images.
Row (a) are the samples recovered from full 12288 features and the average percent deviation is 1.7226e-04\%, which proves that we can recover the image losslessly from full features.
Row (b) are the samples recovered from 9000 features and the average percent deviation is 5.48\%.
Rows (c) and (d) are the samples recovered from 6000 and 3000 features and the average percent deviations are 12.41\% and 15.51\% respectively.
Observing that as the reduced features increase in this level, the deviations grow higher.
Also in rows (c) and row (d), the reconstructed images contain certain insignificant details which look like noises.
As a result, the reconstruction quality and the testing accuracy are poor due to overfitting as we can see in the previous example in  Table~\ref{pldc}.
When the number of features reduces to 270, however, in row (e), we  see that the  insignificant details disappear and the effect just like passing the images through a smoothing process.
Consequently, the deviation is only 9.11\% which is even lower than row (c) and row (d).
This result also reflects that 270 features are suitable for training the classifier since they grasp most of the significant information of an image and avoid overfitting occurred as using 6000 and 3000 features.

\begin{figure}
    \centering
    \includegraphics[width=0.35\textwidth]{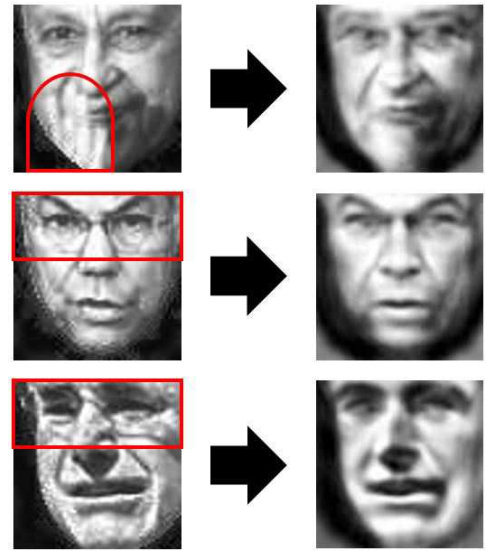}
    \caption{The image recovery method eliminates disturbing objects. The top image originally has fingers in front of the man's chin and disappears after recovered. The middle and the bottom images both have glasses at first but disappear after recovered.}
    \label{recovers2}
\end{figure}

Moreover, from the top and bottom samples of the first column in Fig. \ref{recovers}, the glasses disappear in the reconstructed image from the reduced 270 features.
This implies that the proposed scheme can remove redundancy that is irrelevant or unimportant to classification.
To see this point more clearly,  Fig.~\ref{recovers2} shows more reconstructed images from the 270 features.
Observed from the figure that glasses and figures disappear.
That is, significant features extracted by the proposed scheme would be the information about face structures, not those disturbing objects like glasses and fingers.
Therefore, when we reconstruct the images from a suitable amount of features, the disturbing items should disappear.
From a viewpoint of data compression, the compressed data is dedicated to achieving better classification performance.
This effect may be treated as ``feature filtering'' of the proposed system in classification.

\textbf{Experiment 5: Comparisons between conventional and proposed schemes.}
In this experiment, we compare the proposed system with the AlexNet and the Saak transform in \cite{saak}.
The same preprocessed training and testing images are used.
We apply the AlexNet provided by Matlab.
The CNN model is trained using exactly the same layers like those in AlexNet.
Moreover, considering the hardware, we allow that the CNN provided by Matlab uses GPU to speed up, and the proposed scheme only uses CPU.
The GPU that CNN uses is  NVIDIA GeForce GTX 1050 Ti.
For the Saak transform, 600 features are used, which can achieve its best accuracy.
The result is shown in Table \ref{vs}.
We see that the proposed model achieves a better accuracy while the whole computational time including training and testing is only around one seven hundredths of that with CNN.
\begin{table*}[!htb]
\centering
\caption{Alexnet vs. Saak transform vs. Proposed recognition model}
\begin{tabular}{|c|c|c|c|c|c|c|c|c|c|}
\hline
& \multicolumn{3}{|c|}{AlexNet} & \multicolumn{3}{|c|}{Saak Transform (600 features)} & \multicolumn{3}{|c|}{Proposed Scheme (270 features)}\\ \hline
$\#$ of  & Time & Time & Accuracy & Time & Time & Accuracy & Time & Time & Accuracy\\
classes & (Train) & (Test) & & (Train) & (Test) & & (Train) & (Test) & \\ \hline
$19$ & $550$s & $4.1$s & $\approx 86\%$ & $218$s & $14.3$s & $87.83\%$ & $0.7$s & $0.06$s & $\mathbf{97.61\%}$\\ \hline
$158$ & $1260$s & $12.6$s & $\approx 63\%$ & $2032$s & $49.3$s & $61.93\%$  & $1.7$s & $0.14$s & $\mathbf{84.91\%}$\\ \hline
\end{tabular}
\label{vs}
\end{table*}

\section{Conclusion and future work}
We have proposed a linear classification system that can inverse the extracted and reduced features to original data, and achieve data compression for classification purposes as well.
Experimental results show that the proposed system outperforms the conventional classification schemes in terms of not only computational complexity but also testing accuracy.
From the viewpoint of data compression, the proposed system compresses the data in a way beneficial for classification purposes.
That is, when the images are recovered from the reduced features via the proposed system, several unimportant feature redundancies for classifications such as glasses and covered hand on the faces are naturally filtered out. We call this effect feature filtering.
When the compressed data is used for classification, the testing accuracy is high while the recovered images can still achieve a small percent deviation as well.
The nice properties including linearity, reversibility, achieving data compression and feature filtering have made the proposed system worth further investigation.
The solutions to develop sophisticated algorithms to refine the detection results among top guesses, and to efficiently utilize the advantages of data compression for classification are still open.

{\large

}

\end{document}